\documentclass{article}

\usepackage{placeins}
\usepackage{longtable,booktabs,array}
\usepackage{ltablex}
\usepackage{adjustbox}
\usepackage{subcaption} 

\usepackage{PRIMEarxiv}
\usepackage{multirow}
\usepackage{tcolorbox}
\usepackage{capt-of}
\usepackage{enumitem}
\usepackage{float}
\usepackage{svg}

\usepackage[utf8]{inputenc} 
\usepackage[T1]{fontenc}    
\usepackage{hyperref}       
\usepackage{url}            
\usepackage{booktabs}       
\usepackage{amsfonts} 
\usepackage{pdflscape}
\usepackage{tabularx}
\usepackage{ragged2e}
\usepackage{rotating}
\usepackage{nicefrac}       
\usepackage{microtype}      
\usepackage{lipsum}
\usepackage{fancyhdr}       
\usepackage{graphicx}   \usepackage{subcaption}
\graphicspath{{media/}}     

\title{MetaLLMix : An XAI Aided LLM-Meta-learning Based Approach for Hyper-parameters Optimization
}

\makeatletter
\renewcommand\@fnsymbol[1]{\ifcase#1\or 1\or 2\or 3\or 4\or 5\or 6\or 7\or 8\else\@ctrerr\fi}
\makeatother

\author{
  Bal-Ghaoui Mohamed \footnotemark[1]{} \\ %
  R\&D Department, Audensiel Conseil \\
  Paris, France \\
  \texttt{m.bal-ghaoui@audensiel.fr} \\
  \And
  Tiouti Mohammed \footnotemark[2]{} \\
  Université Évry Paris-Saclay \\
  Evry, France\\
  \texttt{medtiouti965@gmail.com} \\
}

\begin{document}
\maketitle

\vspace{-1cm}
\begin{center}
    11 September 2025\\
    $^1$ Corresponding author: research idea, methodology, review, validation. \\
    $^2$ Contributing author: software development, AI implementation, writing. \\
\end{center}
\vspace{0.5cm}

\begin{abstract}
Effective model and hyperparameter selection remain one of the most challenging aspects of deep learning development, traditionally demanding extensive domain expertise and computational resources through iterative experimentation. While AutoML frameworks and recently large language models (LLMs) have emerged as promising solutions for automating this process, existing LLM-based approaches still require extensive trial-and-error phases, rely on expensive industrial APIs, lack generalizability, and provide limited interpretability through chain-of-thought reasoning or training history without quality control mechanisms.
We propose MetaLLMiX, a novel zero-shot hyperparameter optimization framework that combines meta-learning, explainable AI, and efficient LLM reasoning. Our approach leverages historical experiment outcomes enriched with SHAP-based explanations to simultaneously recommend optimal hyperparameter configurations and pretrained model selections without requiring additional trials.
We also employ LLM-as-judge evaluation approach to assess and control the format, accuracy, and completeness of the generated outputs.
Experimental evaluation on eight diverse medical imaging datasets using nine open-source lightweight LLMs demonstrated that MetaLLMiX achieved competitive to superior performance compared to traditional HPO methods while significantly reducing the computational cost. Unlike previous work that relied solely on commercial APIs, our local deployment approach achieves optimal results on 5 out of 8 classification tasks and demonstrates consistently competitive performance across all remaining datasets, with an optimization time reduced from hours to seconds (99.6-99.9\% reduction in response time). Additionally, MetaLLMiX achieved the fastest training times on 6 out of 8 datasets, with configurations training 2.4x to 15.7x faster than traditional methods, while maintaining accuracy within the 1-5\% of best-performing baselines when not optimal.
\end{abstract}


\section{Introduction}
The field of deep learning has experienced remarkable progress across diverse domains, including computer vision, natural language processing, and medical imaging. However, selecting appropriate model architectures and tuning hyperparameters remains a challenging and computationally intensive process. Traditional hyperparameter optimization (HPO) methods such as Grid Search (GS), Random Search (RS), or Bayesian Optimization (BO), while effective, often lack the ability to generalize across datasets and require extensive computational resources~\cite{franceschi2024hpo,bischl2021foundations}.

Meta-learning approaches have emerged as a promising solution to reduce computational overhead by leveraging knowledge from related tasks. Transfer HPO methods build on this foundation by reusing hyperparameter configurations from similar datasets, avoiding the need to restart optimization from scratch~\cite{wistuba2021few,vanschoren2018meta}. These approaches rely on dataset meta-features to identify similarities and enable effective knowledge transfer, with systems like Auto-sklearn demonstrating successful warm-starting of HPO using meta-learning techniques~\cite{feurer2022auto}.

The emergence of Automated Machine Learning (AutoML) has addressed some of these challenges by providing end-to-end automation for machine learning workflows. Notable AutoML systems like Auto-Sklearn~\cite{feurer2022auto}, Auto-Gluon \cite{erickson2020autogluon}, and cloud-hosted services such as Amazon SageMaker HPO \cite{perrone2020amazon} and Google Vizier \cite{golovin2017google} have made advanced optimization techniques more accessible to practitioners~\cite{yu2020hpo}. These systems integrate algorithm selection with hyperparameter optimization, addressing the Combined Algorithm Selection and Hyperparameter optimization (CASH) problem \cite{thornton2013auto}, which involves jointly selecting the optimal algorithm and its hyperparameters for a dataset, through comprehensive automation frameworks.

Large language models (LLMs) have demonstrated remarkable capabilities across diverse domains, and automated decision-making tasks such as neural architecture search (NAS) \cite{zheng2023can}, HPO, and Machine Learning (ML) workflows~\cite{yang2023large}. Their success stems from their ability to generalize through in-context learning, leveraging problem descriptions and few-shot examples to adapt to new tasks without fine-tuning~\cite{brown2020language,zhao2023survey}. This flexibility has spurred interest in applying LLMs to optimization problems, where their iterative reasoning aligns naturally with sequential processes like BO.

Recent work explores LLMs as alternatives to traditional HPO methods. OPRO~\cite{yang2023large} for instance employs LLMs as meta-optimizers through iterative prompting, while AgentHPO~\cite{liu2024largeBO} introduces a dual-agent system (Creator and Executor) to automate tuning with improved interpretability. Zhang et al.~\cite{zhang2023using} demonstrate that LLMs using zero-shot initialization and few-shot updates can match or surpass RS and BO under constrained budgets, while LLAMBO~\cite{liu2024large} reimagines BO by replacing its components (e.g., surrogate models, samplers) with specialized LLMs, outperforming traditional methods in regret scores measuring the difference between the best-found solution and the true optimum on benchmarks such as Bayesmark \cite{turnerbayesmark} and HPOBench \cite{eggensperger2021hpobench}. These approaches highlight LLMs' potential to reduce reliance on random initialization and dynamically balance exploration-exploitation trade-offs, as further explored by SLLMBO~\cite{mahammadli2024sequential} which integrates LLMs with BO techniques through adaptive search spaces and an LLM-based Tree-Structured Parzen Estimator (LLM-TPE) to lower computational costs.


Simultaneously, the integration of LLMs with explainable AI (XAI) has emerged as a promising direction to explain ML algorithms' decisions. Although this potential  has been  explored in prior research ~\cite{kroeger2024incontextexplainersharnessingllms, 2023burtonxaiml,bhattacharjee2024llmguidedcausalexplainabilityblackbox}, their incorporation into HPO workflows remains largely unexplored. Explingo~\cite{zytek2024explingo} for instance introduced systems that convert SHAP (SHapley Additive exPlanations)\cite{lundberg2017unified} outputs into textual explanations using LLMs, demonstrating the potential for human-interpretable insights. HPExplorer~\cite{grushetskaya-2024} on the other hand, provided semi-automated explainable HPO exploration, and Explainable Bayesian Optimization (XBO) approaches like TNTRules~\cite{chakraborty2024explainable} that generate rule-based explanations for optimization decisions. These techniques remain challenging to interpret for non-experts due to their technical complexity.

Current research efforts in LLM-powered HPO remain constrained by several critical issues. Most methods still require extensive trial-and-error phases and iterative chat histories, which fail to escape the computational burden of traditional optimization approaches. They also depend heavily on proprietary models (e.g., GPT-4) with large parameter counts, leaving cost-effective open-source alternatives largely unexplored. In addition, token consumption costs tend to scale poorly with optimization complexity \cite{yang2023large}. Finally, interpretability mechanisms rely primarily on Chain-of-Thought (CoT)~\cite{wei2022chain} reasoning or simple trial summaries, which provide unreliable efficiency and lack quality control over recommendations. Furthermore, these approaches typically focus solely on hyperparameter tuning while neglecting the equally important problem of model selection.

To address these limitations, we present MetaLLMix (Meta-Learning based Large Language Model for Explainable HPO), a novel zero-shot hyperparameter and model selection recommendation system that combines meta-learning, LLMs, and XAI techniques. We study this framework in a transfer learning setup for medical image classification, leveraging knowledge transfer across related medical imaging tasks.

MetaLLMix addresses key challenges in HPO through three innovations: (1) exploiting meta-learning knowledge from related medical imaging past hyperparameter configurations performances, (2) utilizing LLMs for reasoning and decision-making, and (3) incorporating SHAP values for interpretable explanations of hyperparameter influence. Our framework operates in a zero-shot setting using a relatively small parameter LLMs, making it suitable for resource-constrained environments while providing transparent explanations for hyperparameter recommendations.

The main contributions of this work include:
\begin{enumerate}
    \item A zero-shot optimization approach that eliminates iterative search phases by directly inferring competitive configurations from meta-learning knowledge in a single step, addressing the computational overhead of current LLM-based HPO methods requiring multiple cycles of LLM inference, and performance evaluation feedback

    \item Joint LLM-based optimization of both hyperparameters and model architecture selection within a unified framework

    \item Efficient utilization of smaller, open-source LLMs (less than 8B parameters) achieving comparable performance to larger commercial models, reducing development and computational cost

    \item SHAP-driven natural language explanations that provide quantitative justification while enhancing the transparency and trustworthiness of automated decisions
    
    \item LLM-as-a-judge  evaluation mechanism to control hyperparameter configuration and explanation output format, accuracy, and completeness
    \item Construction of a comprehensive meta-dataset from medical image classification tasks enriched with metadata and high performing configurations
\end{enumerate}
\FloatBarrier

This work aims to facilitate the wider use of deep learning by providing a HPO method that is accessible, efficient, and interpretable for professionals in various fields. In this work, we demonstrate our approach using medical imaging classification tasks through transfer learning, where transparency and reliability are particularly important.

The remainder of this paper is organized as follows: Section~2 presents our methodology, detailing the MetaLLMix framework architecture, meta-dataset construction, and LLM-based reasoning approach. Section~3 describes our experimental setup and presents results from transfer learning experiments across medical image classification tasks. Section~4 discusses the implications of our findings, limitations, and future research directions.

\section{Methodology}

\begin{figure}[htbp] 
  \centering
  \includegraphics[width=1\linewidth]{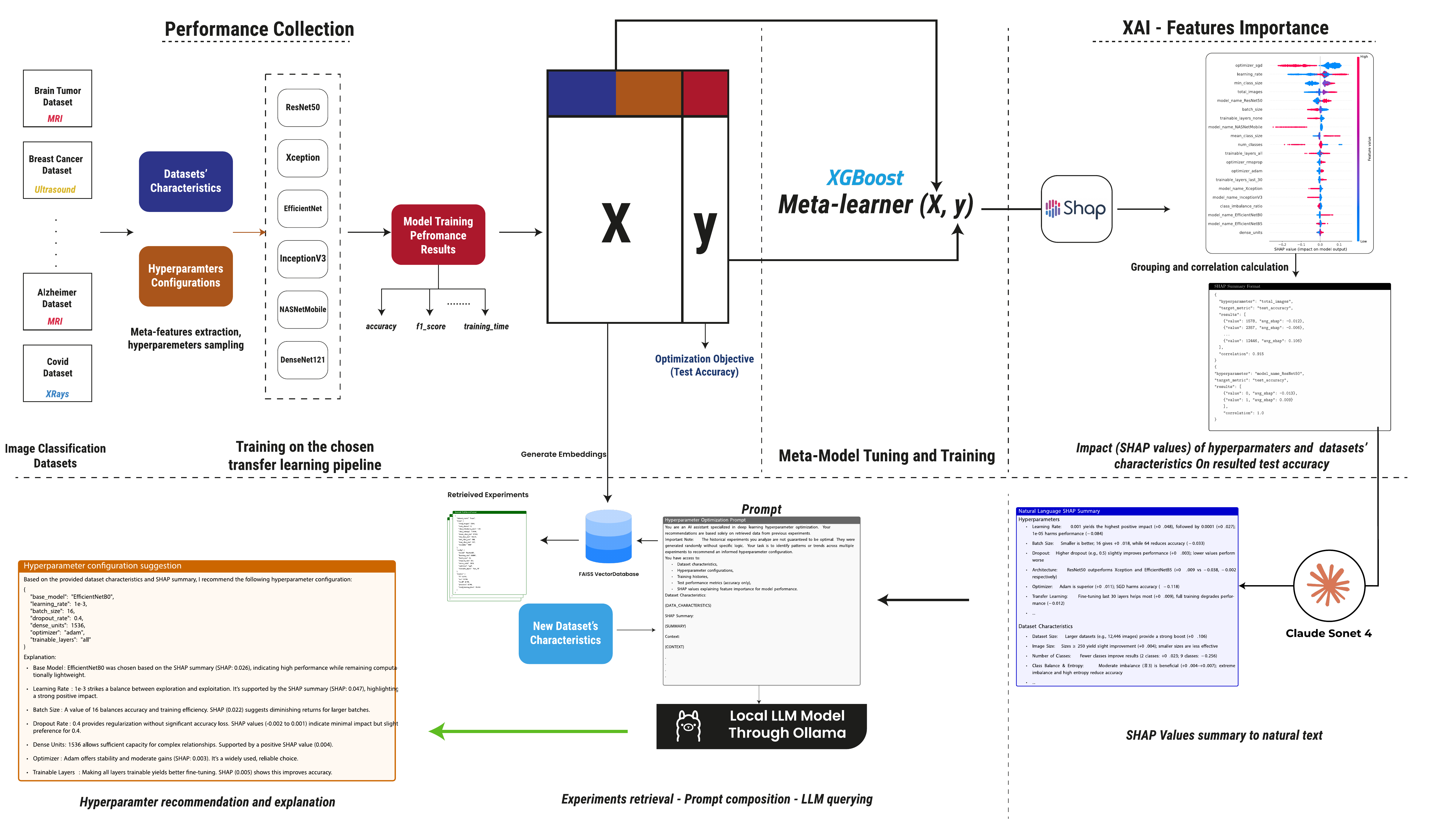}
  \caption{MetaLLMiX full pipeline}
  \label{fig:fig1}
\end{figure}

MetaLLMix is a zero-shot hyperparameter optimization framework that combines meta-learning, XAI, and LLMs to automate model selection and hyperparameter optimization. The full pipeline is demonstrated in Figure~\ref{fig:fig1}. The framework operates through four key phases: (1) meta-dataset construction through comprehensive performance assessment, (2) meta-learner training for performance prediction, (3) SHAP-based explanation generation, and (4) LLM-driven recommendation with natural language justification.
The system leverages historical experimental data from diverse medical imaging datasets to build a comprehensive knowledge base. For each new target dataset, MetaLLMix analyzes dataset characteristics and retrieves similar historical experiments to inform hyperparameter recommendations without requiring iterative optimization.

\subsection{Meta-dataset construction}
To capture the complexity and learning characteristics of each dataset, we extracted seven key meta-features that influence classification performance by analyzing class distribution statistics and dataset structural properties~\cite{rivolli2019characterizingclassificationdatasetsstudy}: number of instances (total count of images), number of classes (count of distinct classification categories), class imbalance ratio (ratio between largest and smallest class sizes), class entropy (Shannon entropy quantifying class distribution evenness), image resolution (pixel dimensions of input images), statistical class descriptors (we chose to include mean, standard deviation, minimum, and maximum class sizes), and imaging modality (medical imaging type such as CT, MRI, X-ray, for which we are studying the effect). These meta-features provide quantitative measures of dataset complexity that facilitate similarity-based retrieval, though their effectiveness for generalization across medical imaging domains requires further validation. 

We implemented a standardized transfer learning pipeline to ensure experimental consistency across all datasets. The architecture comprises seven pre-trained convolutional neural networks (ResNet50\cite{resnet}, Xception\cite{xception}, EfficientNetB5\cite{efficientnet}, EfficientNetB0\cite{efficientnet}, DenseNet121\cite{densenet}, InceptionV3\cite{inception}, NASNetMobile\cite{nasnet}) trained on ImageNet serving as feature extractors, followed by a custom classification head consisting of a Global Average Pooling layer, a dense layer with configurable units, batch normalization with ReLU activation and dropout regularization, and a final dense output layer with softmax activation. The selected models cover a range of complexities and diverse architectures, with several being lightweight and well-suited for constrained environments, such as EfficientNetB0, NASNetMobile, and DenseNet121, while others, like ResNet50 and EfficientNetB5, are more powerful but heavier, enabling comprehensive analysis of architectural effects on model performance. Furthermore, adopting a transfer learning approach introduces additional hyperparameters, such as fine-tuning strategies, which have shown significant differences in handling medical images \cite{bal2023transfer}.

\begin{figure}[htbp] 
  \centering
  \includegraphics[width=.7\textwidth]{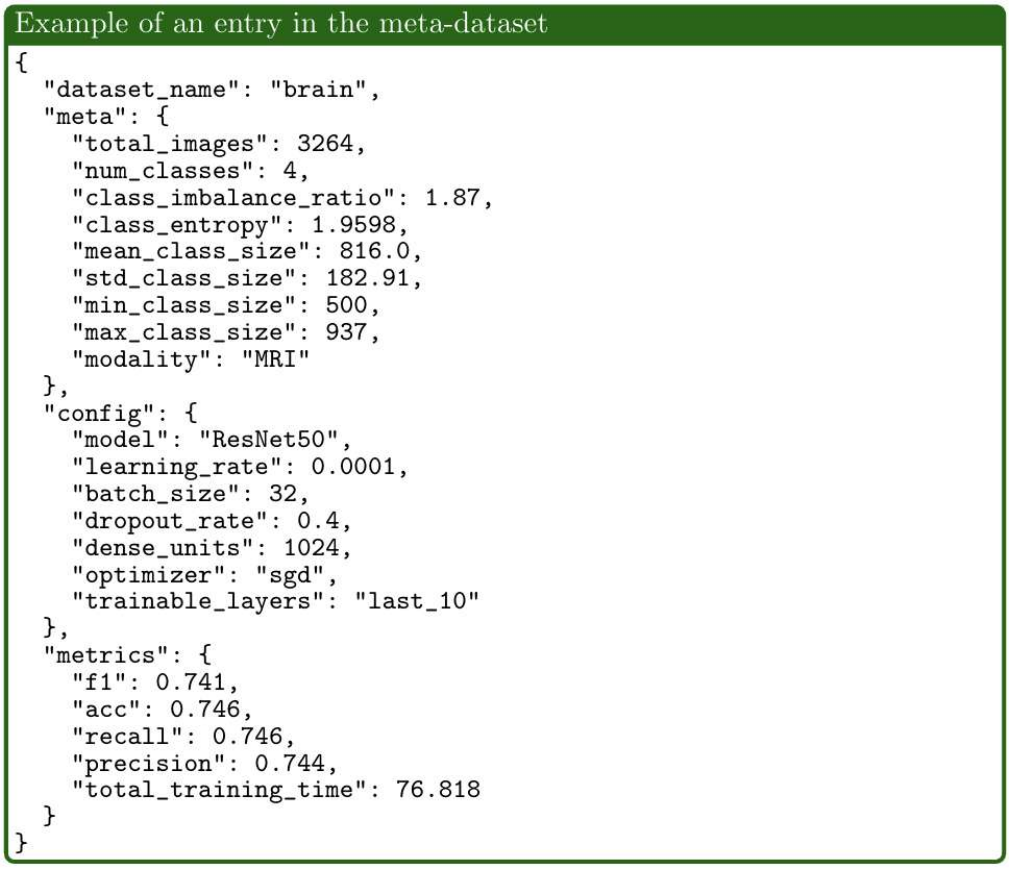}
  \caption{Example of an entry in the meta-dataset}
  \label{fig:entryex}
\end{figure}

Hyperparameter configurations are sampled using RS and BO to ensure diverse coverage of the search space (detailed in Table~\ref{tab:opt_space}), which consists of seven parameters grouped into three categories: base model, classification head parameters, and training parameters. For each dataset-hyperparameter-model combination, we record comprehensive performance metrics including training accuracy, validation accuracy, test accuracy, precision, recall, F1-score, and training time. The resulting meta-dataset combines performance outcomes with dataset meta-features to create training examples for the meta-learner, resulting in entries such as the example shown in Figure \ref{fig:entryex}.

\begin{table}[htbp]
\centering
\caption{Optimization search space explored in MetaLLMix}
\label{tab:opt_space}
\setlength{\tabcolsep}{5pt} 
\begin{tabularx}{\linewidth}{l X p{0.45\linewidth}}
\toprule
\textbf{Category} & \textbf{Parameter} & \textbf{Values} \\
\midrule
\multirow{2}{*}{\textbf{Base model}}
& Base Model & Xception, EfficientNetB5, ResNet50, InceptionV3, NASNetMobile, DenseNet121, EfficientNetB0  \\ \cmidrule{2-3}
& Fine-tuning Strategy & Feature extraction only, Partial fine-tuning (10/30 layers), Full fine-tuning \\
\midrule
\multirow{2}{*}{\textbf{Classification head parameters}}
& Dense Units & 512, 1024, 1536 \\ \cmidrule{2-3}
& Dropout Rate & 0.3, 0.4, 0.5 \\
\midrule
\multirow{3}{*}{\textbf{Training parameters}}
& Optimizer & Adam, SGD, RMSprop \\ \cmidrule{2-3}
& Learning Rate & $10^{-3}$, $10^{-4}$, $10^{-5}$ \\ \cmidrule{2-3}
& Batch Size & 16, 32, 64 \\
\bottomrule
\end{tabularx}
\end{table}

\subsection{Meta-Learner and explainability}

An XGBoost regression model served as the meta-learner, predicting model performance based on dataset meta-features and hyperparameter configurations. This choice aligns with established methodologies in meta-learning, which frequently employ gradient boosting algorithms as meta-learners due to their robust predictive capabilities \cite{woznica2021towards}. The meta-learner's predictive accuracy is crucial as it directly impacts the quality of subsequent SHAP explanations.

We employed SHAP analysis using the TreeSHAP \cite{lundberg2020local} algorithm to generate interpretable explanations for performance predictions (Figure~\ref{fig:prompt}). The SHAP values are systematically grouped by feature values and averaged to reveal global patterns of hyperparameter influence across datasets. Pearson correlation coefficients quantify relationships between feature values and SHAP contributions, providing directional guidance for hyperparameter recommendations.
A dedicated prompting strategy converts SHAP values into human-readable insights through manual interaction with Claude Sonnet 4~\cite{anthropic2025claude}, accessed via the web-based chat interface, utilizing carefully crafted prompts to generate structured interpretations that identify hyperparameters with positive/negative influence, optimal value ranges, and influential dataset characteristics. The processed SHAP results, including averaged contributions and correlation coefficients, are then structured as foundational input for LLM-based reasoning.

\subsection{LLM-based recommendation system}

We employed open-source lightweight LLMs as the core decision-making components through Ollama \cite{ollama}. These smaller models provide analytical reasoning capabilities while maintaining computational efficiency, making the framework accessible in resource-constrained environments.

\begin{figure}[htbp] 
  \centering
  \includegraphics[width=0.7\textwidth]{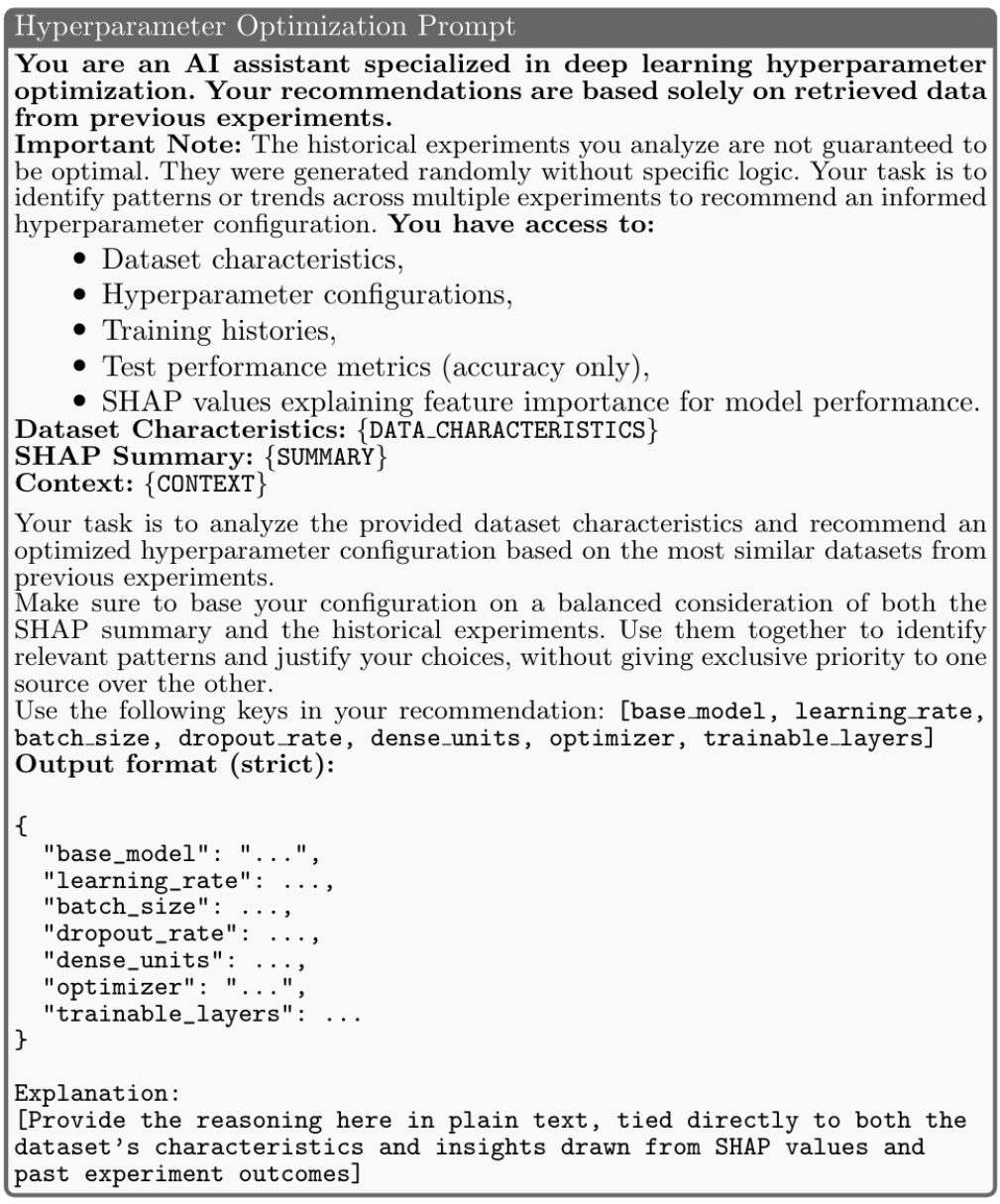}
  \caption{Hyperparameter optimization prompt template}
  \label{fig:prompt}
\end{figure}

The prompt design underwent extensive iterative refinement to ensure reliable and interpretable recommendations (Figure~\ref{fig:prompt}). Key components include a core instruction that establishes the LLM's role as a specialized optimization assistant, constraining recommendations to empirical evidence from historical experiments; context clarification that addresses the limitation that historical experiments were generated through random sampling rather than optimal design; structured inputs that process three information types simultaneously (dataset meta-features, historical experiment outcomes with performance metrics, and SHAP value summaries explaining feature importance); and an output format that enforces strict JSON format for hyperparameter recommendations followed by natural language explanations.

To enhance reasoning capabilities and reduce hallucinations, we integrated a Retrieval Augmented Generation (RAG) \cite{lewis2020retrieval} mechanism that retrieves relevant historical experiments. The system uses FAISS (Facebook AI Similarity Search) \cite{douze2024faiss} for efficient nearest-neighbor search over meta-feature representations, retrieving the top-8 most similar experiments, to provide sufficient diverse historical experiments while fitting within different LLMs' context windows without truncation or slowing down inference time, for each new dataset using LangChain \cite{langchain}. Each retrieved experiment includes dataset metadata (image count, class distribution metrics, and modality information), configuration parameters (complete hyperparameter specifications), performance metrics (evaluation results including accuracy, precision, recall, F1-score, training time), and local SHAP explanations (three most influential features ranked by absolute SHAP value magnitude). This retrieval approach provides concrete historical evidence for LLM reasoning while maintaining interpretability and avoiding the need for dedicated model training over large-scale HPO logs.

\subsection{Recommendation generation}

\begin{figure}[htbp] 
    \centering
    \includegraphics[width=1\linewidth]{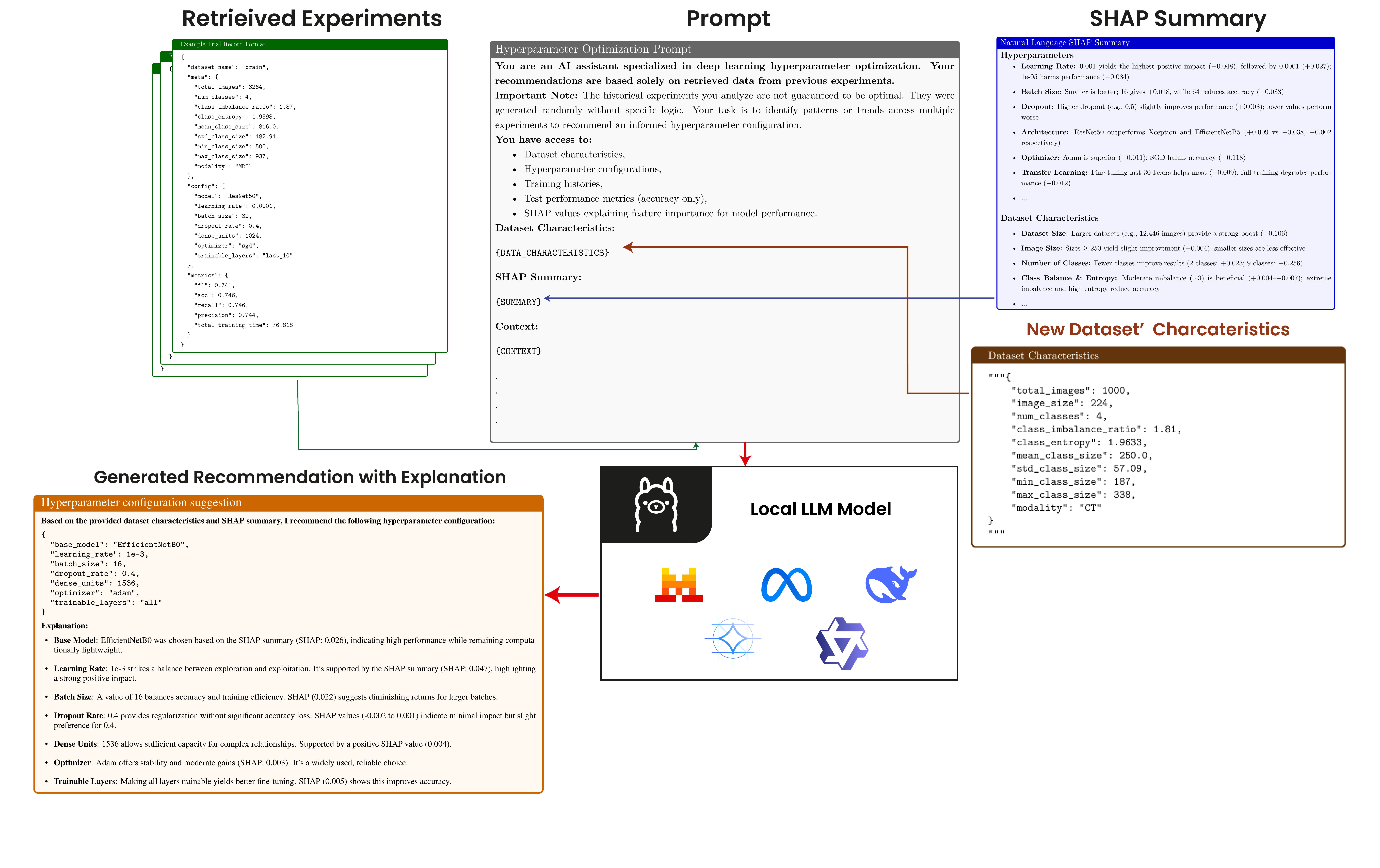}
    \caption{Quering the LLM}
    \label{fig:query}
\end{figure}

To summarize, the complete MetaLLMix pipeline operates through three sequential steps: first, given a new dataset, meta-features are computed, by analyzing class frequency distributions and extracting structural dataset properties, preparing the historical experiment database to be queried using FAISS similarity search; second, retrieved experiments, SHAP summaries, and dataset characteristics are combined into a structured prompt, the assembled context is then processed through the language model to generate hyperparameter recommendations as the final step in a JSON-formatted output, along with a natural language explanation justifying the choices (Figure~\ref{fig:query}). This integrated approach enables zero-shot hyperparameter optimization while providing transparent, evidence-based explanations for each recommendation.  Building upon \cite{zytek2024explingo}, which assessed LLM explanations of ML model decisions, the generated LLM outputs are further evaluated by a separate LLM. In this context, the evaluation focuses on hyperparameter optimization and model selection. It assigns a score based on whether the output format complies with the prompt instructions and whether the explanation accurately incorporates the provided assets (dataset characteristics, SHAP summaries, and history logs). The evaluation further measures the completeness of the explanation (ensuring all relevant inputs are addressed) and its conciseness.

\section{Results}

\subsection{Experimental setup and metrics}
To evaluate the performance of our proposed hyperparameter optimization method, MetaLLMiX, we conducted extensive experiments on eight diverse medical imaging classification tasks. Our experimental setup was designed to evaluate both the predictive performance and the practical utility of the method.

We compared MetaLLMiX against two established state-of-the-art methods: BO with a Gaussian process surrogate model implementation from scikit-optimize, and RS, with both methods executed for 20 iterations. Additionally, we evaluated nine different open-source LLMs to assess how they influence performance and interpretability.

Using 4-fold cross-validation technique, three key metrics were computed including: test accuracy, optimization response time (time required to generate hyperparameter configurations), and training time (duration to train the final model with the selected configurations). Additionally, LLM output evaluation was performed through a judge LLM attributing scores 
ranging from 0 to 4 measuring output’s accuracy, completeness, conciseness, fluency, and conformity to specified output formatting. These scores were obtained by running each assessment 3 times and taking the mean to account for potential variability in LLMs’ responses.

\subsection{Datasets}
\begin{figure}[htbp]
    \centering
    \includegraphics[width=1\linewidth]{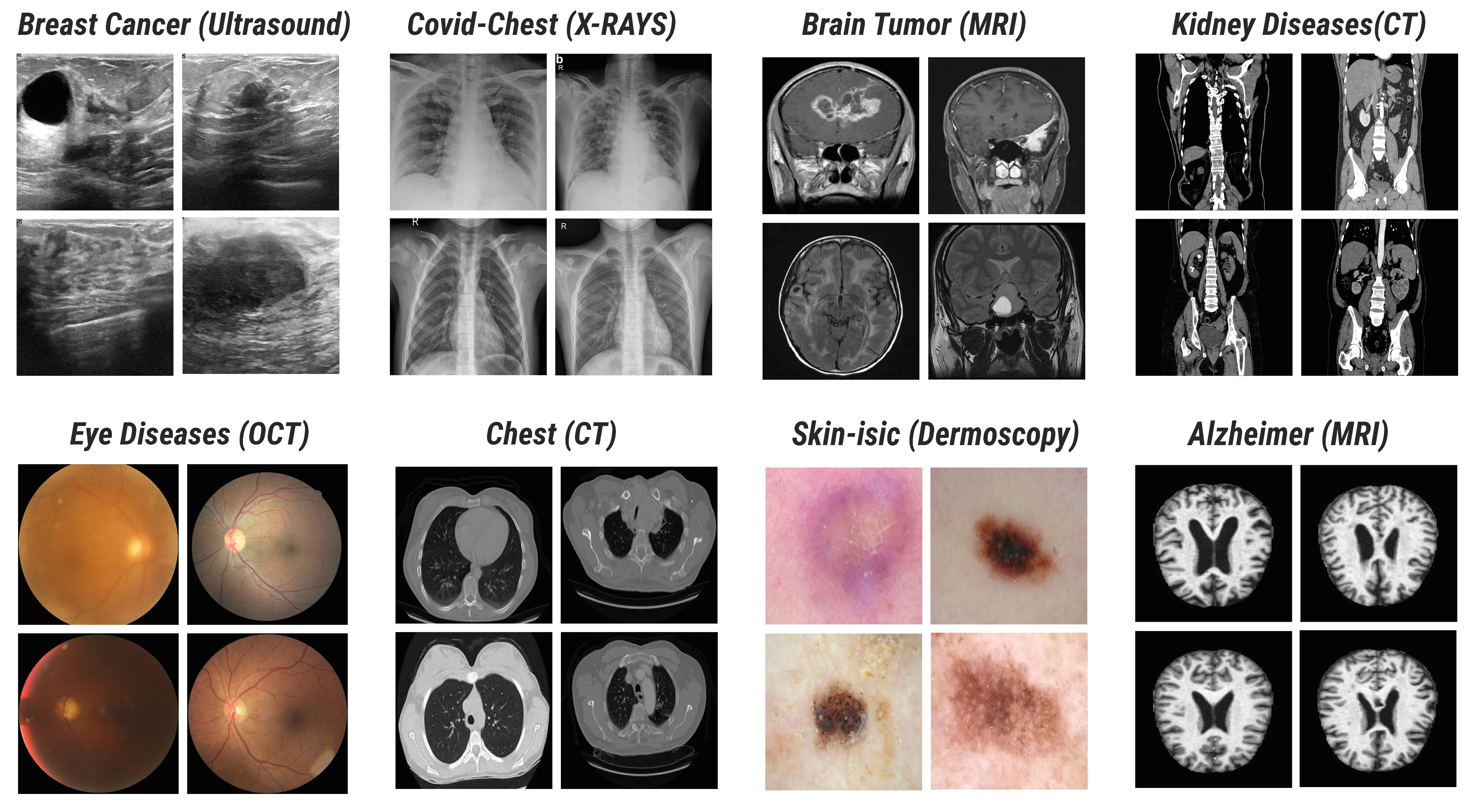}
    \caption{Medical Imaging Classification Datasets}
    \label{fig:medicaldatasets}
\end{figure}

We employed a diverse collection of publicly available medical imaging datasets to validate the generalizability of MetaLLMiX, as detailed in Table \ref{tab:datasets}. These datasets cover various imaging modalities such as MRI, CT, X-ray, US, DMS, and OCT, and address diagnostic tasks including brain tumors, Alzheimer’s disease, COVID-19 detection, and dermatological conditions (Figure \ref{fig:medicaldatasets}).

This dataset diversity ensures that the evaluation of MetaLLMiX reflects a wide range of medical imaging scenarios and allows us to assess its robustness across domains.

\setcounter{table}{1} 
\begin{table}[H] 
\centering
\caption{Summary of datasets used for Meta-learning}
\label{tab:datasets}
\begin{tabularx}{\textwidth}{l >{\RaggedRight}X ccc} 
\toprule
\textbf{Dataset} & \textbf{Description} & \textbf{Modality} & \textbf{\# Images} & \textbf{\# Classes} \\
\midrule
Alzheimer~\cite{tourist55alzheimers} & MRI images of brain labeled across dementia stages & MRI & 6,400 & 4 \\
\addlinespace
Brain Tumor~\cite{sartaj_bhuvaji_ankita_kadam_prajakta_bhumkar_sameer_dedge_swati_kanchan_2020} & MRI scans for classifying benign, malignant, pituitary tumors, etc. & MRI & 3,264 & 4 \\
\addlinespace
Breast Cancer~\cite{al-dhabyani-2019} & Ultrasound scans classified as normal, benign, and malignant & US & 780 & 3 \\
\addlinespace
Chest CT~\cite{chestCTscan} & CT scans labeled as normal or with 3 types of lung cancer & CT & 1,000 & 4 \\
\addlinespace
Chest COVID X-Rays~\cite{cohen2020covidProspective} & Chest X-rays with COVID-19, SARS, ARDS, pneumonia, normal. & X-ray & 5,910 & 2 \\
\addlinespace
Eye Disease OCT~\cite{eyeOCT} & OCT images showing diabetic retinopathy, glaucoma, cataract, etc. & OCT & 4,217 & 4 \\
\addlinespace
Kidney Diseases~\cite{kidneyCT} & CT scans of kidneys with tumor, cyst, stone, or normal diagnosis & CT & 12,446 & 4 \\
\addlinespace
Skin - ISIC~\cite{skinISIC} & Dermoscopic skin lesion images labeled by various skin conditions & DMS & 2,357 & 9 \\
\bottomrule
\end{tabularx}
\end{table}

\subsection{Performance comparison with traditional HPO methods}

\begin{figure}[ht]
    \centering

    \begin{subfigure}{1\textwidth}
        \centering
        \includegraphics[width=\linewidth]{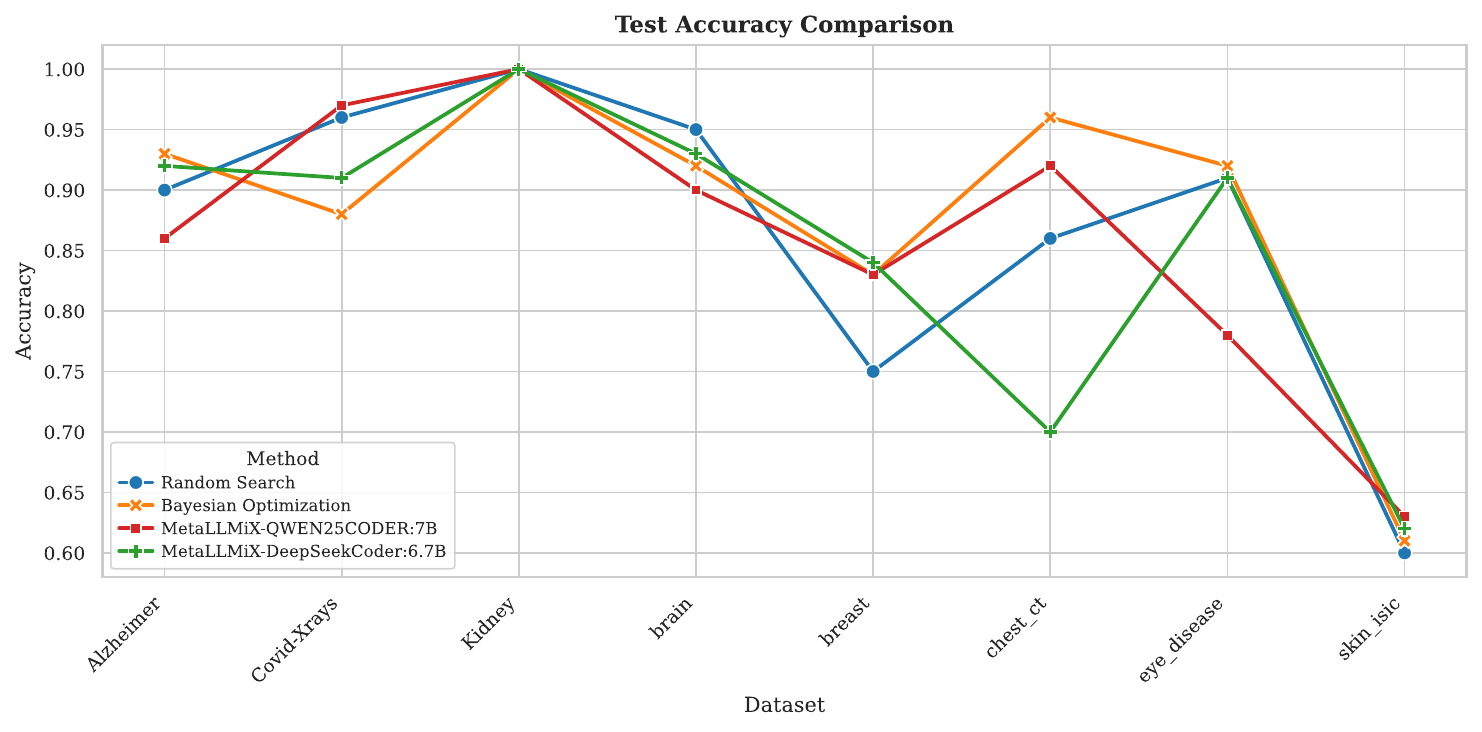}
        \caption{Test Accuracy}
    \end{subfigure}

    \vspace{0.1em} 

    \begin{subfigure}{0.8\textwidth}
        \centering
        \includegraphics[width=\linewidth]{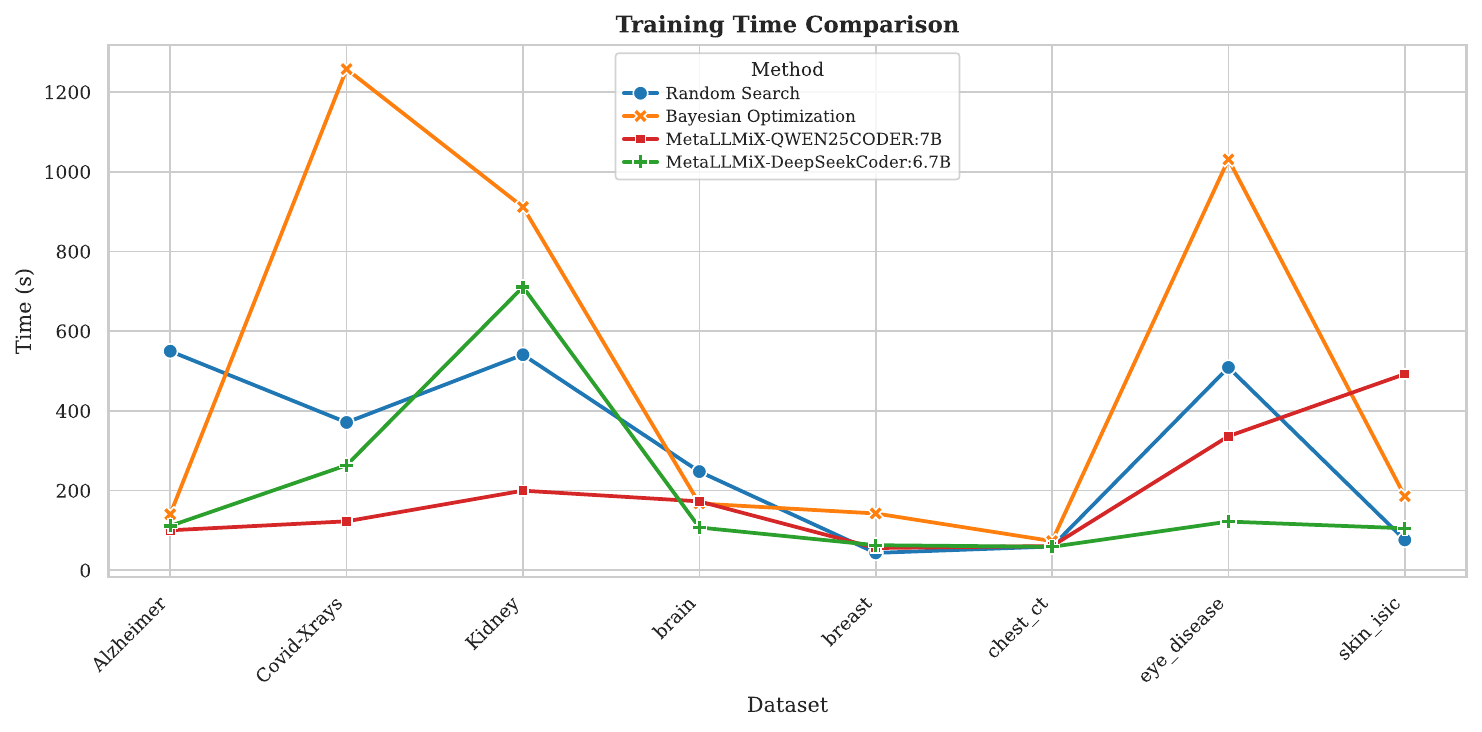}
        \caption{Training Time}
    \end{subfigure}


    \begin{subfigure}{0.8\textwidth}
        \centering
        \includegraphics[width=\linewidth]{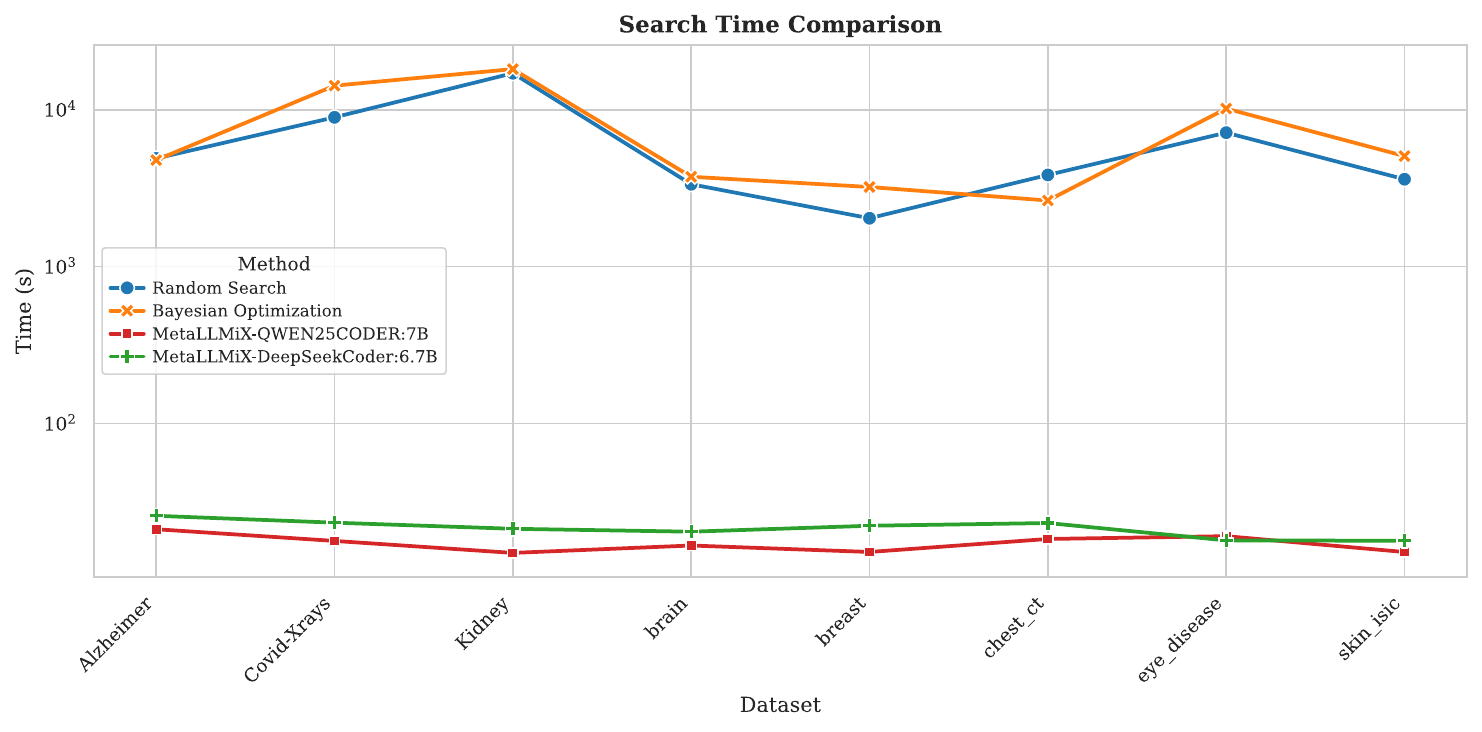}
        \caption{Search Time}
    \end{subfigure}

    \caption{MetaLLMix and classic HPO evaluation results across datasets}
    \label{fig:hpo_comparison_grid}
\end{figure}

\begin{figure}[ht]
    \centering

    \begin{subfigure}{\textwidth}
        \begin{minipage}{0.15\textwidth} 
            \caption{Test Accuracy}
        \end{minipage}%
        \begin{minipage}{0.75\textwidth} 
            \includegraphics[width=\linewidth]{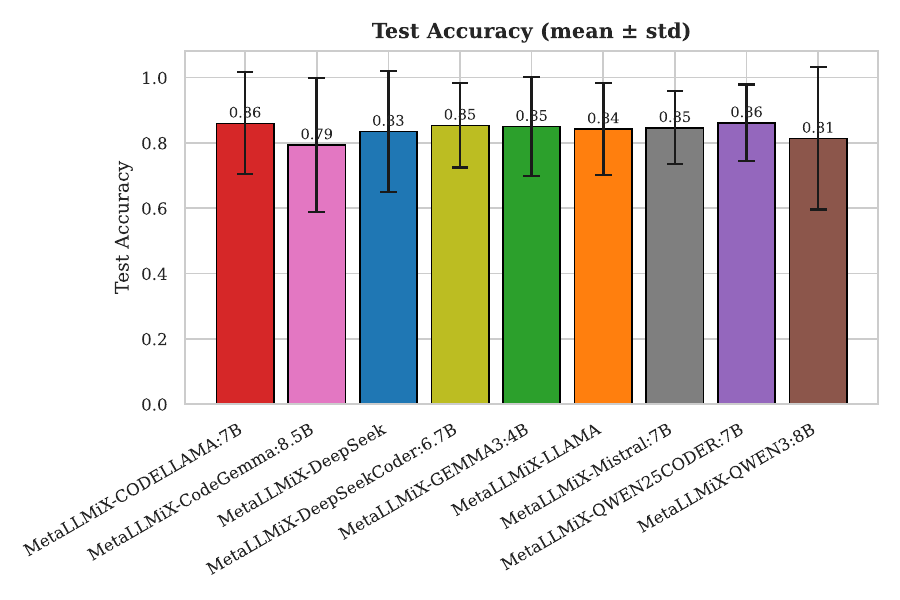}
        \end{minipage}
    \end{subfigure}

    \vspace{-1em}

    \begin{subfigure}{\textwidth}
        \begin{minipage}{0.15\textwidth}
            \caption{Search Time}
        \end{minipage}%
        \begin{minipage}{0.75\textwidth}
            \includegraphics[width=\linewidth]{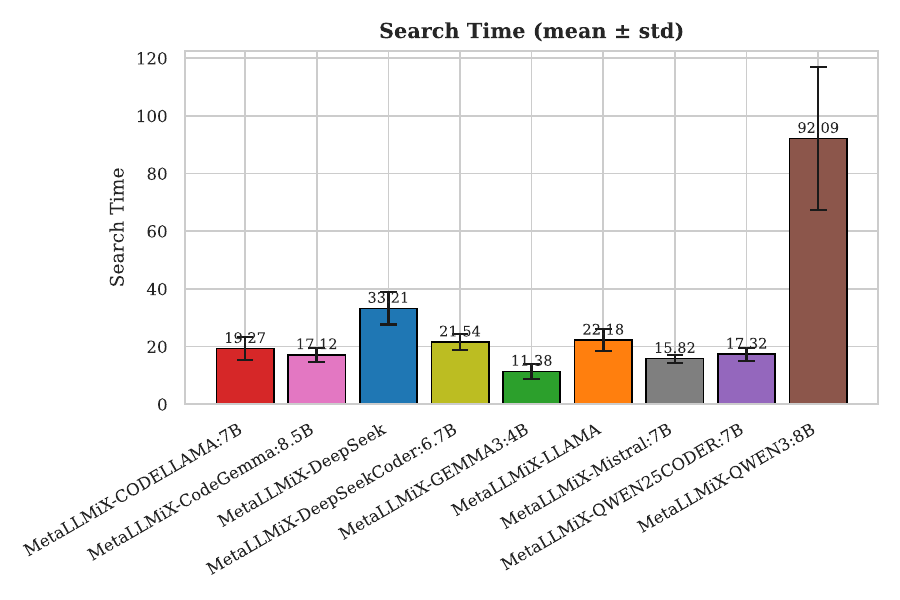}
        \end{minipage}
    \end{subfigure}


    \begin{subfigure}{0.49\textwidth}
        \centering
        \includegraphics[width=\linewidth]{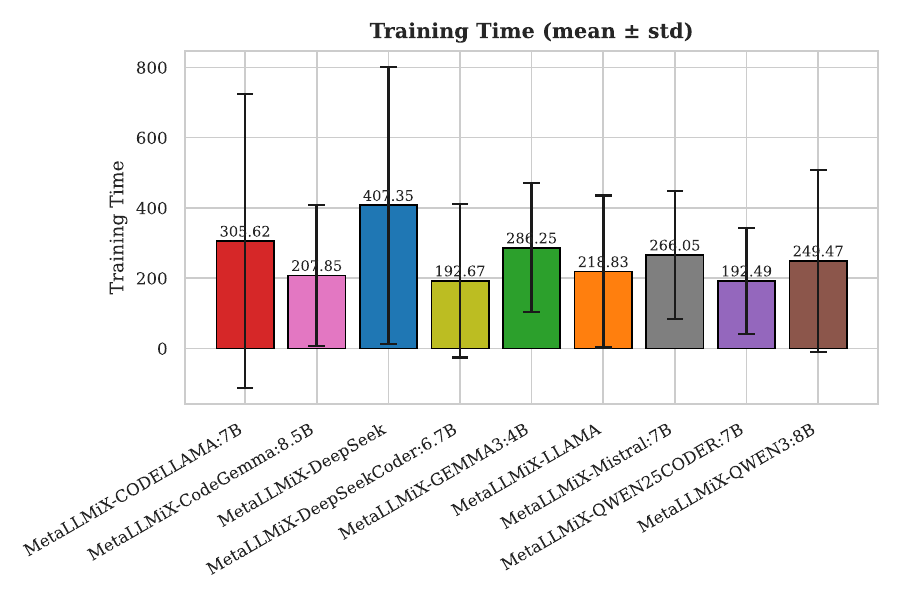}
        \caption{Training Time}
    \end{subfigure}
    \begin{subfigure}{0.49\textwidth}
        \centering
        \includegraphics[width=\linewidth]{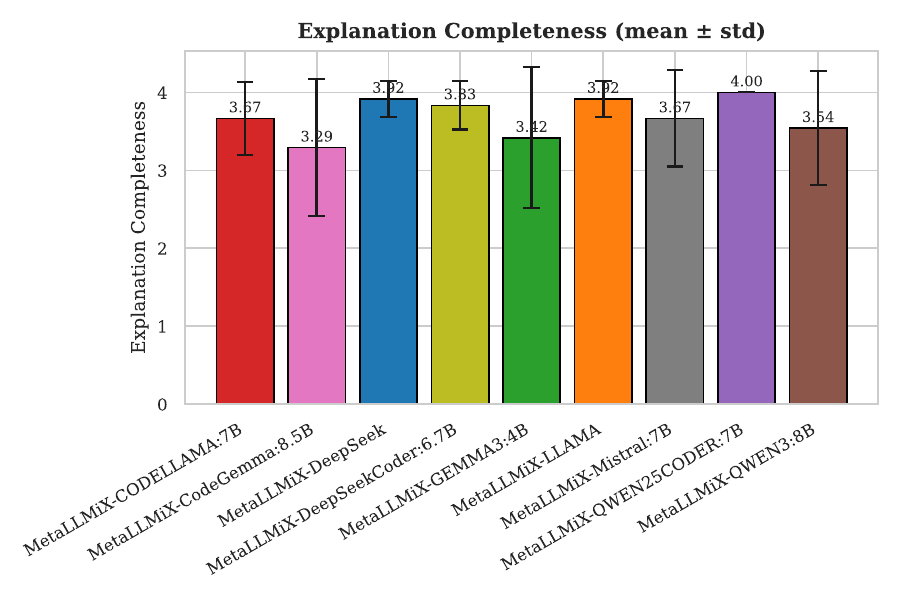}
        \caption{Explanation Completeness}
    \end{subfigure}

    \caption{Language model comparison across all datasets}
    \label{fig:llm_comparison_grid}
\end{figure}

Table~\ref{tab:hpo_comparison} reports the evaluation results of our proposed MetaLLMiX framework in comparison with RS and BO, using its optimal LLM configuration to demonstrate the framework’s effectiveness with an appropriate model selection. Figures~\ref{fig:hpo_comparison_grid} and~\ref{fig:llm_comparison_grid} illustrate the performance of the MetaLLMix approach using different LLMs across multiple datasets.

To assess the impact of different language models on MetaLLMiX performance, we evaluated nine open-source LLMs across all medical imaging datasets presented previously. The comparative results are reported in Table \ref{tab:llm_analysis}.

\vspace{0.5cm}
\setcounter{table}{2}
\begin{table}[H] 
\centering
\caption{MetaLLMix comparison results against classic HPO methods}
\label{tab:hpo_comparison}
\begin{tabular}{llccc}
\toprule
\textbf{Dataset} & \textbf{Method} & \textbf{CV Accuracy} & \textbf{Response Time (s)} & \textbf{Training Time (s)} \\
\midrule
\multirow{3}{*}{Alzheimer}
  & Random Search & 0.90 & 4893.66 & 549.59 \\
  & Bayesian Opt. & \textbf{0.93} & 4773.95 & 140.45 \\
  & MetaLLMiX (Ours) & \textbf{0.92} & \textbf{12.79} & \textbf{88.19} \\
\midrule
\multirow{3}{*}{Brain}
  & Random Search & \textbf{0.95} & 3348.67 & 247.44 \\
  & Bayesian Opt. & 0.92 & 3738.41 & 167.43 \\
  & MetaLLMiX (Ours) & \textbf{0.95} & \textbf{13.51} & \textbf{73.98} \\
\midrule
\multirow{3}{*}{Breast}
  & Random Search & 0.75 & 2033.66 & \textbf{43.40} \\
  & Bayesian Opt. & 0.83 & 3216.13 & 142.37 \\
  & MetaLLMiX (Ours) & \textbf{0.84} & \textbf{10.31} & \textbf{43.71} \\
\midrule
\multirow{3}{*}{Chest Ct}
  & Random Search & 0.86 & 3837.86 & 59.47 \\
  & Bayesian Opt. & \textbf{0.96} & 2637.79 & 73.06 \\
  & MetaLLMiX (Ours) & \textbf{0.92} & \textbf{7.55} & \textbf{53.41} \\
\midrule
\multirow{3}{*}{Covid-Xrays}
  & Random Search & 0.96 & 8942.73 & 370.78 \\
  & Bayesian Opt. & 0.88 & 14232.60 & 1257.37 \\
  & MetaLLMiX (Ours) & \textbf{0.97} & \textbf{13.13} & \textbf{122.69} \\
\midrule
\multirow{3}{*}{Eye Disease}
  & Random Search & 0.91 & 7139.25 & 508.86 \\
  & Bayesian Opt. & \textbf{0.92} & 10164.52 & 1030.84 \\
  & MetaLLMiX (Ours) & \textbf{0.91} & \textbf{10.72} & \textbf{122.06} \\
\midrule
\multirow{3}{*}{Kidney}
  & Random Search & 1.00 & 17085.13 & 540.74 \\
  & Bayesian Opt. & 1.00 & 18145.48 & 911.27 \\
  & MetaLLMiX (Ours) & \textbf{1.00} & \textbf{9.35} & \textbf{199.61} \\
\midrule
\multirow{3}{*}{Skin Isic}
  & Random Search & 0.60 & 3604.04 & \textbf{76.00} \\
  & Bayesian Opt. & 0.61 & 5064.34 & 185.66 \\
  & MetaLLMiX (Ours) & \textbf{0.63} & \textbf{11.27} & \textbf{105.19} \\
\bottomrule
\end{tabular}
\end{table}

\keepXColumns 
\begin{small}     
\begin{tabularx}{\textwidth}{l l l c c c c c c c c c}
\caption{Comparison results of language models\label{tab:llm_analysis}} \\

\toprule
\textbf{Dataset} & \textbf{Method} & \textbf{Sel. Model} & \textbf{Acc} & \textbf{R.T (s)} & \textbf{T.T (s)} & \textbf{Comp.} & \textbf{Flu.} & \textbf{Conc.} & \textbf{Cons.} & \textbf{Form.} & \textbf{Rank} \\
\midrule
\endfirsthead

\multicolumn{12}{l}{\small\textit{Table \thetable\ continued from previous page}}\\
\toprule
\textbf{Dataset} & \textbf{Method} & \textbf{Sel. Model} & \textbf{Acc} & \textbf{R.T (s)} & \textbf{T.T (s)} & \textbf{Comp.} & \textbf{Flu.} & \textbf{Conc.} & \textbf{Cons.} & \textbf{Form.} & \textbf{Rank} \\
\midrule
\endhead

\bottomrule
\multicolumn{12}{r}{\small\textit{Continued on next page}} \\
\endfoot

\bottomrule
\endlastfoot

\multirow{9}{*}{Alzheimer}
 & deepseek-r1:7b & ResNet50 & 0.75 & 33.50 & 164.02 & 4 & 4 & 4 & 4 & 4 & 9 \\
 & llama3.1:8b & EfficientNetB0 & 0.69 & 23.60 & \textbf{88.19} & 4 & 4 & 4 & 4 & 4 & 8 \\
 & gemma3:4b & EfficientNetB5 & 0.77 & \textbf{12.79} & 234.14 & 4 & 4 & 4 & 4 & 4 & 6 \\
 & codellama:7b & EfficientNetB5 & 0.89 & 15.41 & 103.78 & 4 & 4 & 4 & 3 & 4 & 2 \\
 & qwen2.5-coder:7b & ResNet50 & 0.86 & 21.19 & 100.18 & 4 & 4 & 4 & 4 & 4 & 4 \\
 & qwen3:8b & ResNet50 & 0.86 & 94.84 & 93.03 & 4 & 4 & 4 & 4 & 4 & 7 \\
 & codegemma:7b & EfficientNetB5 & 0.88 & 18.88 & 100.33 & 4 & 4 & 4 & 3 & 4 & 3 \\
 & mistral:7b & ResNet50 & 0.80 & 15.53 & 275.65 & 4 & 4 & 4 & 4 & 4 & 5 \\
 & \textbf{deepseek-coder:6.7b} & EfficientNetB5 & \textbf{0.92} & 25.81 & 111.23 & 4 & 4 & 3 & 4 & 4 & \textbf{1} \\
\midrule
\multirow{9}{*}{Brain}
 & deepseek-r1:7b & EfficientNetB5 & 0.95 & 33.61 & 271.60 & 4 & 4 & 4 & 4 & 4 & 8 \\
 & llama3.1:8b & ResNet50 & 0.91 & 17.80 & \textbf{73.98} & 3 & 4 & 4 & 3 & 4 & 2 \\
 & gemma3:4b & EfficientNetB5 & 0.91 & \textbf{13.51} & 189.25 & 4 & 4 & 4 & 4 & 4 & 5 \\
 & codellama:7b & ResNet50 & 0.93 & 18.59 & 95.91 & 3 & 2 & 4 & 2 & 4 & 3 \\
 & qwen2.5-coder:7b & ResNet50 & 0.90 & 16.69 & 172.70 & 4 & 4 & 4 & 4 & 4 & 4 \\
 & qwen3:8b & EfficientNetB0 & 0.93 & 88.30 & 106.33 & 3 & 3 & 4 & 3 & 4 & 9 \\
 & codegemma:7b & EfficientNetB5 & 0.79 & 18.94 & 94.54 & 4 & 4 & 4 & 4 & 4 & 6 \\
 & mistral:7b & EfficientNetB5 & 0.81 & 15.27 & 247.24 & 4 & 4 & 4 & 4 & 4 & 7 \\
 & \textbf{deepseek-coder:6.7b} & ResNet50 & \textbf{0.93} & 20.44 & 107.25 & 3 & 4 & 4 & 3 & 4 & \textbf{1} \\
\midrule
\multirow{9}{*}{Breast}
 & deepseek-r1:7b & EfficientNetB0 & 0.81 & 30.56 & 55.76 & 4 & 4 & 4 & 4 & 4 & 5 \\
 & llama3.1:8b & ResNet50 & 0.83 & 19.49 & 54.62 & 4 & 4 & 4 & 3 & 4 & 2 \\
 & gemma3:4b & EfficientNetB5 & 0.82 & \textbf{10.31} & 129.25 & 2 & 3 & 4 & 2 & 4 & 8 \\
 & codellama:7b & EfficientNetB5 & 0.82 & 17.14 & 137.26 & 4 & 4 & 4 & 3 & 4 & 7 \\
 & qwen2.5-coder:7b & EfficientNetB0 & \textbf{0.83} & 15.18 & 55.68 & 4 & 4 & 4 & 3 & 4 & \textbf{1} \\
 & qwen3:8b & ResNet50 & 0.80 & 134.87 & \textbf{43.71} & 4 & 4 & 4 & 4 & 4 & 9 \\
 & codegemma:7b & ResNet50 & \textbf{0.84} & 18.61 & 53.70 & 4 & 4 & 4 & 3 & 4 & 3 \\
 & mistral:7b & EfficientNetB0 & 0.83 & 14.82 & 129.93 & 3 & 4 & 4 & 3 & 4 & 6 \\
 & \textbf{deepseek-coder:6.7b} & EfficientNetB5 & \textbf{0.84} & 22.30 & 62.74 & 4 & 4 & 4 & 4 & 4 & 4 \\
\midrule
\multirow{9}{*}{Chest Ct}
 & deepseek-r1:7b & ResNet50 & 0.91 & 36.21 & 63.89 & 4 & 4 & 4 & 4 & 4 & 5 \\
 & llama3.1:8b & DenseNet121 & 0.90 & 19.32 & 97.06 & 4 & 4 & 4 & 4 & 4 & 4 \\
 & gemma3:4b & EfficientNetB5 & 0.91 & \textbf{7.55} & 162.88 & 2 & 3 & 4 & 2 & 4 & 6 \\
 & codellama:7b & EfficientNetB0 & 0.87 & 17.64 & 62.19 & 4 & 4 & 4 & 4 & 4 & 3 \\
 & \textbf{qwen2.5-coder:7b} & ResNet50 & \textbf{0.92} & 18.37 & 60.65 & 4 & 4 & 4 & 4 & 4 & \textbf{1} \\
 & qwen3:8b & EfficientNetB0 & 0.91 & 61.35 & 63.45 & 4 & 4 & 4 & 4 & 4 & 7 \\
 & codegemma:7b & ResNet50 & 0.45 & 19.48 & 70.50 & 3 & 4 & 4 & 4 & 4 & 9 \\
 & mistral:7b & ResNet50 & 0.89 & 18.15 & \textbf{53.41} & 4 & 4 & 4 & 4 & 4 & 2 \\
 & deepseek-coder:6.7b & EfficientNetB5 & 0.70 & 23.21 & 58.77 & 4 & 4 & 4 & 4 & 4 & 8 \\
\midrule
\multirow{9}{*}{Covid-Xrays}
 & deepseek-r1:7b & EfficientNetB5 & 0.94 & 31.77 & 1243.81 & 4 & 4 & 4 & 4 & 4 & 8 \\
 & llama3.1:8b & EfficientNetB0 & 0.96 & 30.15 & 377.41 & 4 & 4 & 2 & 4 & 4 & 6 \\
 & gemma3:4b & EfficientNetB5 & 0.96 & 15.53 & 389.21 & 4 & 4 & 4 & 4 & 4 & 2 \\
 & codellama:7b & EfficientNetB5 & \textbf{0.97} & 16.53 & 1318.97 & 2 & 4 & 4 & 2 & 4 & 7 \\
 & \textbf{qwen2.5-coder:7b} & EfficientNetB0 & \textbf{0.97} & 17.85 & \textbf{122.69} & 4 & 4 & 4 & 3 & 4 & \textbf{1} \\
 & qwen3:8b & ResNet50 & 0.81 & 72.03 & 727.41 & 4 & 4 & 4 & 4 & 4 & 9 \\
 & codegemma:7b & EfficientNetB0 & \textbf{0.97} & \textbf{13.13} & 142.00 & 2 & 3 & 4 & 2 & 4 & 4 \\
 & mistral:7b & EfficientNetB0 & 0.95 & 14.44 & 390.56 & 4 & 4 & 4 & 4 & 4 & 3 \\
 & deepseek-coder:6.7b & ResNet50 & 0.91 & 23.33 & 263.56 & 4 & 4 & 4 & 4 & 4 & 5 \\
\midrule
\multirow{9}{*}{Eye Disease}
 & deepseek-r1:7b & EfficientNetB5 & 0.90 & 30.25 & 310.61 & 4 & 4 & 4 & 4 & 4 & 7 \\
 & llama3.1:8b & EfficientNetB5 & 0.87 & 24.15 & 184.83 & 4 & 4 & 4 & 3 & 4 & 4 \\
 & gemma3:4b & ResNet50 & 0.91 & \textbf{10.72} & 336.52 & 4 & 4 & 4 & 4 & 4 & 3 \\
 & codellama:7b & EfficientNetB0 & 0.90 & 19.58 & 345.14 & 3 & 4 & 4 & 3 & 4 & 5 \\
 & qwen2.5-coder:7b & EfficientNetB0 & 0.78 & 19.14 & 336.25 & 4 & 4 & 4 & 4 & 4 & 8 \\
 & qwen3:8b & EfficientNetB5 & 0.90 & 71.34 & 123.63 & 4 & 4 & 4 & 4 & 4 & 9 \\
 & codegemma:7b & EfficientNetB0 & \textbf{0.91} & 17.69 & 363.26 & 3 & 4 & 4 & 3 & 4 & 6 \\
 & mistral:7b & ResNet50 & 0.86 & 14.36 & 192.76 & 4 & 4 & 4 & 4 & 4 & 2 \\
 & \textbf{deepseek-coder:6.7b} & EfficientNetB5 & \textbf{0.91} & 17.99 & \textbf{122.06} & 4 & 4 & 4 & 3 & 4 & \textbf{1} \\
\midrule
\multirow{9}{*}{Kidney}
 & deepseek-r1:7b & ResNet50 & \textbf{1.00} & 44.47 & 627.27 & 3 & 4 & 4 & 4 & 4 & 8 \\
 & llama3.1:8b & EfficientNetB5 & \textbf{1.00} & 22.07 & 688.18 & 4 & 4 & 4 & 4 & 4 & 5 \\
 & gemma3:4b & EfficientNetB5 & \textbf{1.00} & \textbf{9.35} & 682.39 & 4 & 4 & 4 & 4 & 4 & 3 \\
 & codellama:7b & ResNet50 & \textbf{1.00} & 28.43 & 202.68 & 4 & 4 & 3 & 4 & 4 & 2 \\
 & \textbf{qwen2.5-coder:7b} & EfficientNetB0 & \textbf{1.00} & 14.97 & \textbf{199.61} & 4 & 4 & 4 & 4 & 4 & \textbf{1} \\
 & qwen3:8b & EfficientNetB0 & \textbf{1.00} & 97.49 & 571.66 & 3 & 3 & 4 & 3 & 3.333 & 9 \\
 & codegemma:7b & ResNet50 & \textbf{1.00} & 13.72 & 642.09 & 2 & 2 & 4 & 2 & 4 & 7 \\
 & mistral:7b & EfficientNetB5 & \textbf{1.00} & 16.71 & 643.69 & 3 & 4 & 4 & 3 & 4 & 4 \\
 & deepseek-coder:6.7b & EfficientNetB0 & \textbf{1.00} & 21.33 & 710.59 & 3 & 4 & 4 & 4 & 4 & 6 \\
\midrule
\multirow{9}{*}{Skin Isic}
 & deepseek-r1:7b & ResNet50 & 0.42 & 25.31 & 521.82 & 4 & 4 & 4 & 3 & 4 & 8 \\
 & llama3.1:8b & EfficientNetB0 & 0.58 & 20.85 & 186.35 & 4 & 4 & 4 & 3 & 4 & 3 \\
 & gemma3:4b & EfficientNetB5 & 0.52 & \textbf{11.27} & 166.33 & 3 & 4 & 4 & 4 & 4 & 5 \\
 & codellama:7b & EfficientNetB5 & 0.50 & 20.87 & 179.07 & 3 & 4 & 4 & 4 & 4 & 7 \\
 & qwen2.5-coder:7b & EfficientNetB0 & \textbf{0.63} & 15.17 & 492.12 & 4 & 4 & 4 & 4 & 4 & 4 \\
 & qwen3:8b & ResNet50 & 0.30 & 116.54 & 266.56 & 2 & 4 & 3 & 3 & 4 & 9 \\
 & codegemma:7b & EfficientNetB0 & 0.51 & 16.50 & 196.41 & 4 & 4 & 4 & 3 & 4 & 6 \\
 & mistral:7b & EfficientNetB5 & \textbf{0.63} & 17.27 & 195.14 & 3 & 4 & 4 & 4 & 4 & 2 \\
 & \textbf{deepseek-coder:6.7b} & EfficientNetB0 & 0.62 & 17.89 & \textbf{105.19} & 4 & 4 & 4 & 4 & 4 & \textbf{1} \\

\bottomrule
\end{tabularx}

\footnotesize\raggedright
Sel. Model : Selected Model, R.T: Response Time, T.T: Training Time, Comp.: Completeness, Flu.: Fluency, Conc.: Conciseness, Cons.: Consistency, Form.: Format, Rank: Overall Rank.
\end{small}

\FloatBarrier
\section{Discussion}

\subsection{Performance comparison: MetaLLMiX vs traditional HPO methods}

The comparative evaluation demonstrates that MetaLLMiX achieves competitive predictive performance while providing substantial computational efficiency gains over traditional HPO methods. As shown in Table~\ref{tab:hpo_comparison}, MetaLLMiX achieved optimal accuracy on 4 out of 8 datasets (Brain, Breast, Covid-Xrays, and Skin-Isic), matching the performance of traditional methods on the Kidney dataset (100\% accuracy across all approaches), and showing competitive results on the remaining datasets (Figure~\ref{fig:hpo_comparison_grid}).

Notably, MetaLLMiX consistently outperformed both baselines in computational efficiency metrics. The framework achieved response times ranging from 7.55 to 13.51 seconds across all datasets, representing a significant reduction of 99.6-99.9\% compared to RS (2,033-17,085 seconds) and BO (2,637-18,145 seconds). This efficiency gain stems from the zero-shot nature of our approach, which eliminates the iterative search phases required by traditional methods.
Training time analysis reveals that MetaLLMiX tends to recommend lightweight model configurations, resulting in faster training despite this not being explicitly specified in the prompt instructions. The framework achieved the fastest training times on 6 out of 8 datasets, with MetaLLMiX-recommended configurations training 2.4x to 15.7x faster than those suggested by traditional methods across most datasets. Particularly notable reductions were observed on the Alzheimer dataset (88.19s vs. 549.59s for RS configurations) and Covid-Xrays (122.69s vs. 370.78s for RS configurations).
The results reveal that MetaLLMiX maintains predictive accuracy within 1-5\% of the best-performing traditional method on datasets where it does not achieve optimal performance (Alzheimer: 0.92 vs. 0.93 BO; Chest CT: 0.92 vs. 0.96 BO; Eye Disease: 0.91 vs. 0.92 BO). This marginal accuracy trade-off is offset by the substantial computational savings, making MetaLLMiX particularly suitable for resource-constrained environments or rapid prototyping scenarios.
It should be noted that these MetaLLMiX results represent the best-performing LLM configuration from our framework, demonstrating the optimal potential of the approach when paired with appropriate language model selection.

\subsection{Language model performance analysis}
The evaluation across nine open-source LLMs reveals significant variability in MetaLLMiX performance, highlighting the critical importance of appropriate language model selection for hyperparameter optimization tasks. The results revealed significant variability in MetaLLMiX both predictive accuracy and computational efficiency across different LLMs, highlighting the critical importance of an appropriate language model selection for hyperparameter optimization tasks (Figure~\ref{fig:llm_comparison_grid}).

\textbf{Performance variability and inconsistencies:} The analysis reveals considerable inconsistencies in LLM performance across datasets. For instance, on the Skin-ISIC dataset, accuracy ranges from 0.30 (qwen3:8b) to 0.63 (qwen2.5-coder:7b and mistral:7b), representing a substantial performance gap, though this is justifiable given the task difficulty also observed with BO and RS in the presented results. Similarly, on the Chest CT dataset, codegemma:7b achieves only 0.45 accuracy while qwen2.5-coder:7b reaches 0.92. These variations suggest that certain LLMs struggle with specific dataset characteristics or fail to properly interpret the given context.

\textbf{Model size and performance relationship:} The results challenge the assumption that larger models consistently perform better. The 4B parameter gemma3 model achieves competitive accuracy across most datasets while maintaining the fastest response times (7.55-15.53 seconds). Conversely, the 8B parameter models show mixed results, with llama3.1:8b and qwen3:8b demonstrating inconsistent performance patterns and notably slower response times for qwen3:8b (61.35-134.87 seconds).

\textbf{Explanation quality assessment:} Judge LLM scores reveal varying explanation quality across models. Most LLMs achieve perfect scores (4.0) for completeness, fluency, and conciseness on the majority of datasets. However, notable exceptions include gemma3:4b and codegemma:7b, which show reduced completeness scores (2.0-3.33) on specific datasets, and several models demonstrating lower consistency scores, particularly codellama:7b (2.0-2.67 on multiple datasets).

\textbf{Computational efficiency patterns:} Response time analysis shows that model architecture significantly impacts inference speed. Gemma3:4b consistently achieves the fastest response times across 6 out of 8 datasets, while qwen3:8b demonstrates the slowest performance with response times exceeding 60 seconds on most tasks. This efficiency-accuracy trade-off suggests that smaller, well-optimized models may be preferable for practical deployments.

\textbf{Dataset-specific sensitivities:} Certain datasets appear more challenging for LLM-based optimization. The Skin-ISIC dataset shows the highest performance variability, with multiple models failing to achieve reasonable accuracy levels. This suggests that complex multi-class problems with subtle feature distinctions may be more difficult for LLMs to optimize effectively through meta-learning approaches. Furthermore, the absence of similar datasets in the meta-dataset seems to impair knowledge transfer, as observed also with the Alzheimer dataset, where no comparably imbalanced dataset was present to guide the optimization effectively.

These findings underscore the need for careful LLM selection in automated hyperparameter optimization systems, as model performance cannot be predicted solely from general language modeling capabilities or specialized training domains. Moreover, the results highlight the importance of extending and diversifying the meta-dataset to include a wider range of dataset characteristics, such as highly imbalanced and complex multi-class problems, to improve knowledge transfer and enhance the robustness of LLM-based optimization approaches.

\subsection{Impact of SHAP-Driven explanations on model selection}

\begin{figure}[htbp] 
  \centering
  \includegraphics[width=.7\textwidth]{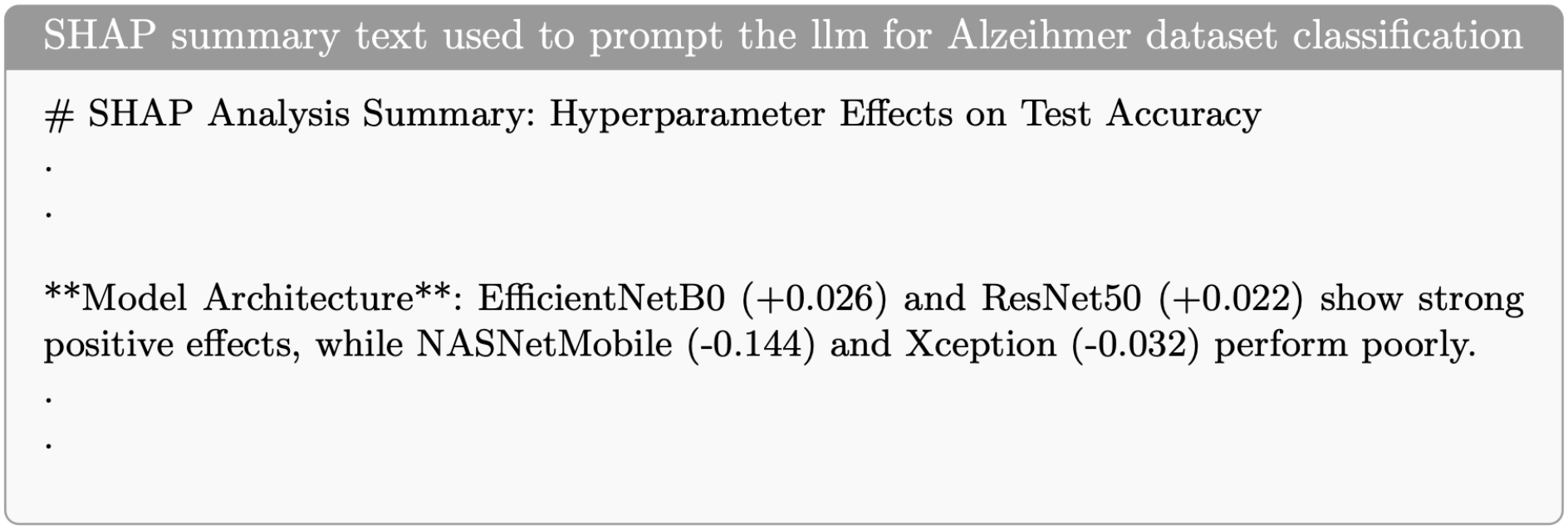}
  \caption{Example of SHAP summary}
  \label{fig:shapsummary}
\end{figure}

The integration of SHAP values provides crucial interpretability insights that guide both hyperparameter and model architecture selection. To illustrate this mechanism, we examine the model selection patterns observed in the Alzheimer dataset, where SHAP analysis revealed distinct preferences for specific architectures. The SHAP summary for model architecture selection on the Alzheimer dataset (Figure~\ref{fig:shapsummary}) demonstrated clear directional preferences.

This quantitative analysis directly influenced LLM decision-making, leading to consistent selection of EfficientNetB0 and ResNet50 architectures across multiple LLM evaluations while avoiding the poorly-performing alternatives.

The SHAP-driven approach provides several advantages over traditional model selection methods. First, it offers quantitative justification for architectural choices rather than relying on heuristics or prior assumptions. Second, the explanations enable practitioners to understand why specific models are recommended for particular dataset characteristics. Finally, the integration of SHAP values with LLM reasoning creates a transparent decision-making process that can be audited and validated by domain experts.

\subsection{Current limitations and drawbacks}

The significant performance variability across different LLMs represents a critical dependency that requires domain-specific evaluation for optimal results. The framework's effectiveness is fundamentally constrained by the quality and diversity of the meta-dataset, with historical experiments generated through random sampling potentially introducing suboptimal patterns.

The current evaluation scope, limited to medical imaging classification tasks, raises questions about generalizability to other domains and tasks. Additionally, certain datasets demonstrate higher sensitivity to LLM selection.

\subsection{Future research directions}

Several promising avenues emerge for extending and improving MetaLLMiX capabilities:

\textbf{Domain expansion and generalization:} Future work should explore MetaLLMiX performance across diverse domains beyond medical imaging. Integration with comprehensive meta-datasets like Meta-Album \cite{ullah2022meta}, which includes 40 datasets from ecology, manufacturing, human actions, and optical character recognition, would provide robust evaluation of transfer learning capabilities across different visual domains. Extension to tabular data, natural language processing, and time-series analysis would demonstrate broader applicability.

\textbf{LLM architecture optimization:} Investigating specialized LLM architectures or fine-tuning approaches specifically designed for hyperparameter optimization could address the performance variability observed across different models. This could include developing domain-specific LLMs trained on HPO tasks or implementing mixture-of-experts approaches that route different optimization problems to specialized model components.

\textbf{Multi-Objective and constrained optimization:} Extending the framework to handle multi-objective scenarios, such as balancing accuracy, training time, computational cost, and model interpretability simultaneously, would increase practical utility. Integration of resource constraints and deployment requirements could make recommendations more practically relevant.

\textbf{Uncertainty quantification and robustness:} Incorporating uncertainty estimation in hyperparameter recommendations would provide practitioners with confidence intervals and risk assessments for optimization decisions. 

\subsection{Conclusion}

MetaLLMiX represents an early exploration of LLM-assisted optimization that provides a zero-shot model and hyperparameter selection without iterative computational burden. The integration of explainable AI through SHAP analysis sets a precedent for transparent, interpretable decision that could influence broader AutoML development. With the ongoing progress in LLM capabilities, the framework could evolve to perform optimization more autonomously, leveraging its enhanced reasoning skills and relying less on the meta-dataset and past experimental results quality.

The successful demonstration of competitive performance using lightweight, open-source models suggests potential for edge deployment    scenarios where optimization decisions must be made locally without external API dependencies. This capability could prove particularly valuable in resource-constrained or privacy-sensitive applications. However, the fundamental challenge of knowledge transfer across domains remains a critical research area. While MetaLLMiX demonstrates promising results within medical imaging, achieving reliable generalization across diverse problem types will require continued advancement in both meta-learning techniques and LLM reasoning capabilities.

\section*{Declarations}
\subsection*{Acknowledgements}
The authors gratefully acknowledge Lab’Innov at Audensiel Conseil R\&D Department for providing the research environment and technical support that enabled this work. This study forms part of the advancement of the internal research and development project STAAC.


\bibliographystyle{unsrt}  
\bibliography{main}  
\end{document}